\documentclass[review]{elsarticle}

\usepackage{lineno,hyperref}
\usepackage{graphicx}
\usepackage[]{algorithm2e}
\RestyleAlgo{ruled}
\usepackage{algpseudocode}
\usepackage{amsmath}
\usepackage{float}
\modulolinenumbers[200]

\journal{arXiv}

\begin{document}

\begin{frontmatter}

\title{Click prediction boosting via Bayesian hyperparameter optimization based ensemble learning pipelines}


\author[first_institute,second_institute]{\c{C}a\u{g}atay Demirel\corref{mycorrespondingauthor}}


\author[third_institute]{A. Aylin Toku\c{c}}
\author[first_institute]{Ahmet Tezcan Tekin}

\address[first_institute]{Computer Engineering Department, Istanbul Technical University, Istanbul, Turkey}
\address[second_institute]{Donders Institute for Brain, Cognition and Behaviour, Radboud University Medical Center, Nijmegen, Netherlands}
\address[third_institute]{Computer Engineering Department, Kadir Has University, Istanbul, Turkey University}

\begin{abstract}
Online travel agencies (OTA's) advertise their website offers on meta-search bidding engines. The problem of predicting the number of clicks a hotel would receive for a given bid amount is an important step in the management of an OTA's advertisement campaign on a meta-search engine, because $bid \times number of clicks$ defines the cost to be generated. Various regressors are ensembled in this work to improve click prediction performance. Following the preprocessing procedures, the feature set is divided into train and test groups depending on the logging date of the samples. The data collection is then subjected to feature elimination via utilizing XGBoost, which significantly reduces the dimension of features. The optimum hyper-parameters are then found by applying Bayesian hyperparameter optimization to XGBoost, LightGBM, and SGD models. The different trained models are tested separately as well as combined to form ensemble models. Four alternative ensemble solutions have been suggested. The same test set is used to test both individual and ensemble models, and the results of 46 model combinations demonstrate that stack ensemble models yield the desired $R^2$ score of all. In conclusion, the ensemble model improves the prediction performance by about 10\%.
\end{abstract}

\begin{keyword}
\texttt{Machine Learning \sep Ensemble Learning \sep Bayesian Optimization
\sep Meta-search Engines \sep Online travel agencies}
\end{keyword}

\end{frontmatter}

\linenumbers

\section{Introduction}

Millions of travelers book hotel accommodation over the Internet each year. Modern travelers rely on peer options, electronic word of mouth (eWOM), and peer reviews. Popular online travel websites offer reliable reviews and prices \cite{bb1}. Therefore, customers choose to inspect and compare different options on meta-search sites like Kayak.com, Trivago, and TripAdvisor before booking their accommodations.

Online travel agencies (OTA's) advertise their website offers on meta-search bidding engines. If the OTA chooses to have a Cost-Per-Click (CPC) ad campaign, the OTA promises to pay a certain amount for each click a certain hotel gets from the platform under predefined conditions. The amount to pay per click is the OTA's $bid$ amount. The problem of predicting the number of clicks a hotel would get for a certain bid amount is an important step in the OTA's advertisement campaign management on a meta-search engine, as $bid \times number of clicks$ defines the cost to be generated.

In one study, state-of-the-art prediction algorithms and Extreme Gradient Boosting (XGBoost) \cite{bb2} regressor as well as a minimum Redundancy-Maximum Relevance (mRMR) \cite{bb3} feature selection algorithm were executed to predict the daily clicks to be received per hotel, using a large OTA's data from Turkey \cite{bb3.1}. The data set received from the meta-search bidding engine contained both numerical and categorical features, with each column having missing and outlier values. The number of clicks as the multiplication of the predicted click-through rate (CTR) and the predicted hotel impression were modelled. The highest R-Squared values obtained in the prediction of individual-hotel based CTR and impression values are both achieved by XGBoost in this work. 

Another study aimed to forecast how many impressions and clicks a hotel will acquire as well as how many rooms it will sell via a meta-search bidding engine \cite{bb4}. The given model predicts how much money an OTA's hotels will make the following day. The authors demonstrate that by incorporating OTA-specific information into prediction models, the generalization of models improves and better results are obtained. In that study, the best results were obtained using tree-based boosting techniques.

Predicting hotel searches, clicks, and bookings is a challenging task due to many external factors such as seasonality, events, location, and hotel-based properties. Capturing such properties increases the accuracy of prediction models. Due to the high variance in daily OTA data, the use of non-linear prediction methods and creating relevant features with a time-delayed data preprocessing approach is adopted in a work trying to forecast daily room sales for each hotel in a meta-search bidding platform \cite{bb5}. They applied XGBoost, random forest, gradient boosting, deep neural networks, and generalized linear models (GLM) \cite{bb5.1}. The most successful model to predict bookings is gradient boosting, applied on a dataset enriched by features that can summarize the trends in the target variable well.

The demand for hotel rooms in the hotel industry in Turkey between the years 2002-2013 is estimated using ARIMA by Efendioglu and Bulkan \cite{bb6}. In their study, they determined the hotel room capacity according to the cost of the unsold rooms and the ARIMA distribution. They also reported that the hotel room demand in the country could be affected by outer factors such as political crises and warnings about terrorism. This work shows the non-deterministic nature of hotel room demand and how unpredictable factors suddenly affect the click prediction problem.

In the literature, studies are focusing on the problem of predicting the CTR of a sponsored display advertisement to be shown on a search engine, related to a query. Click and CTR prediction is an ongoing research for both industry and academia \cite{bb7} \cite{bb8} \cite{bb9}. Our aim of predicting the number of clicks is highly related to the CTR prediction problem, hence those studies are investigated to get a better understanding of related work.

In order to predict ad clicks, Google makes use of logistic regression with improvements in the context of traditional supervised learning based on an FTRL-Proximal online learning algorithm \cite{bb10} for better sparsity and convergence. Microsoft's Bing Search Engine proposes a new Bayesian online learning algorithm for CTR prediction for sponsored search \cite{bb11}, which is based on a probit regression model that maps discrete or real-valued input features to probabilities. The scalability of the algorithm is ensured through a principled weight pruning procedure and an approximate parallel implementation. Yahoo adopts a machine learning framework based on Bayesian logistic regression to predict click-through and conversion rates \cite{bb12}, which is simple, scalable, and efficient. Facebook combines decision trees with logistic regression \cite{bb13}, generating 3\% better results in click prediction, compared to other methods. 

Ensemble learning \cite{bb14} is a machine learning model combination that gets decisions from various models to enhance the overall performance. The ensemble approach provides the stability and low-variety predictions of machine learning algorithms. It builds a set of decision-makers, namely classifiers and regressors, with various techniques as final decisions \cite{bb15}. 

An ensemble model is proposed by Wang et al. to predict the CTR of advertisements on search engines \cite{bb16}. Firstly, they tried several Maximum Likelihood Estimation (MLE)-based methods to exploit the training set; including Online Bayesian Probit Regression (BPR) \cite{bb16.1}, Support Vector Machine (SVM), and Latent Factor Model (LFM) \cite{bb16.2} and optimized them by selecting the most descriptive features. They have created a rank-based ensemble model using the outputs of BPR, SVM, and MLE. The results are ensembled using harmonic means to generate the final blending submission. The proposed model's output shows an on average 0.013 improvement over the individual models.

Ensemble learning techniques implemented by King et al. to investigate whether they could increase the profitability of pay-per-click (PPC) campaigns \cite{bb17}. They applied voting, bootStrap aggregation (Bagging) \cite{bb17.1}, stacked generalization (or stacking) \cite{bb17.2}, and metacost \cite{bb17.3} techniques to four base classifiers, namely, Naïve Bayes, logistic regression, decision trees, and Support Vector Machines. The research in this work analyzed a data set of PPC advertisements placed on the Google search engine, aiming to classify PPC campaign success. They used average accuracy, recall, and precision metrics to measure the performance of both base classifiers and ensemble models. They also introduced the evaluation metric of total campaign portfolio profit and illustrated how relying on overall model accuracy can be misleading. They conclude that applying ensemble learning techniques in PPC marketing campaigns can achieve higher profits. 

Eight ensemble methods were proposed by Ling et al. to accurately estimate the CTR in sponsored search ads \cite{bb18}. A single model would lead to sub-optimal accuracy, and the regression models all have different advantages and disadvantages. The ensemble models are created via bagging, boosting, stacking, and cascading. The training data is collected from historical ads' impressions and the corresponding clicks. The Area under the Receiver Operating Characteristic Curve (AUC) and Relative Information Gain (RIG) metrics are computed against the testing data to evaluate prediction accuracy. They conclude that boosting is better than cascading for the given problem. Boosting neural networks with gradient boosting decision trees turned out to be the best model in the given setting. They conclude that the model ensemble is a promising direction for CTR prediction; meanwhile, domain knowledge is also essential in the ensemble design. 

Etsy, an online e-commerce platform, displays promoted search results, which are similar to sponsored search results and our problem with meta-search bidding engines. CTR prediction is utilized in the system to determine the ranking of the ads \cite{bb19}. They found out that different features capture different aspects, so they classified the features as being historical and content-based. They train separate CTR prediction models based on historical and content-based features, separately. Then, these individual models are combined with a logistic regression model. They reported AUC, Average Impression Log Loss, and Normalized Cross-Entropy metrics to compare the models to non-trivial baselines on a large-scale real-world dataset from Etsy, demonstrating the effectiveness of the proposed system. 

In this study, we utilize ensemble learning pipelines to predict the number of clicks a hotel will receive the next day, and comparing substantial amount of stand-alone prediction performance of the models.


\section{Overview of the Proposed System}

\begin{figure} 
\includegraphics[width=\textwidth]{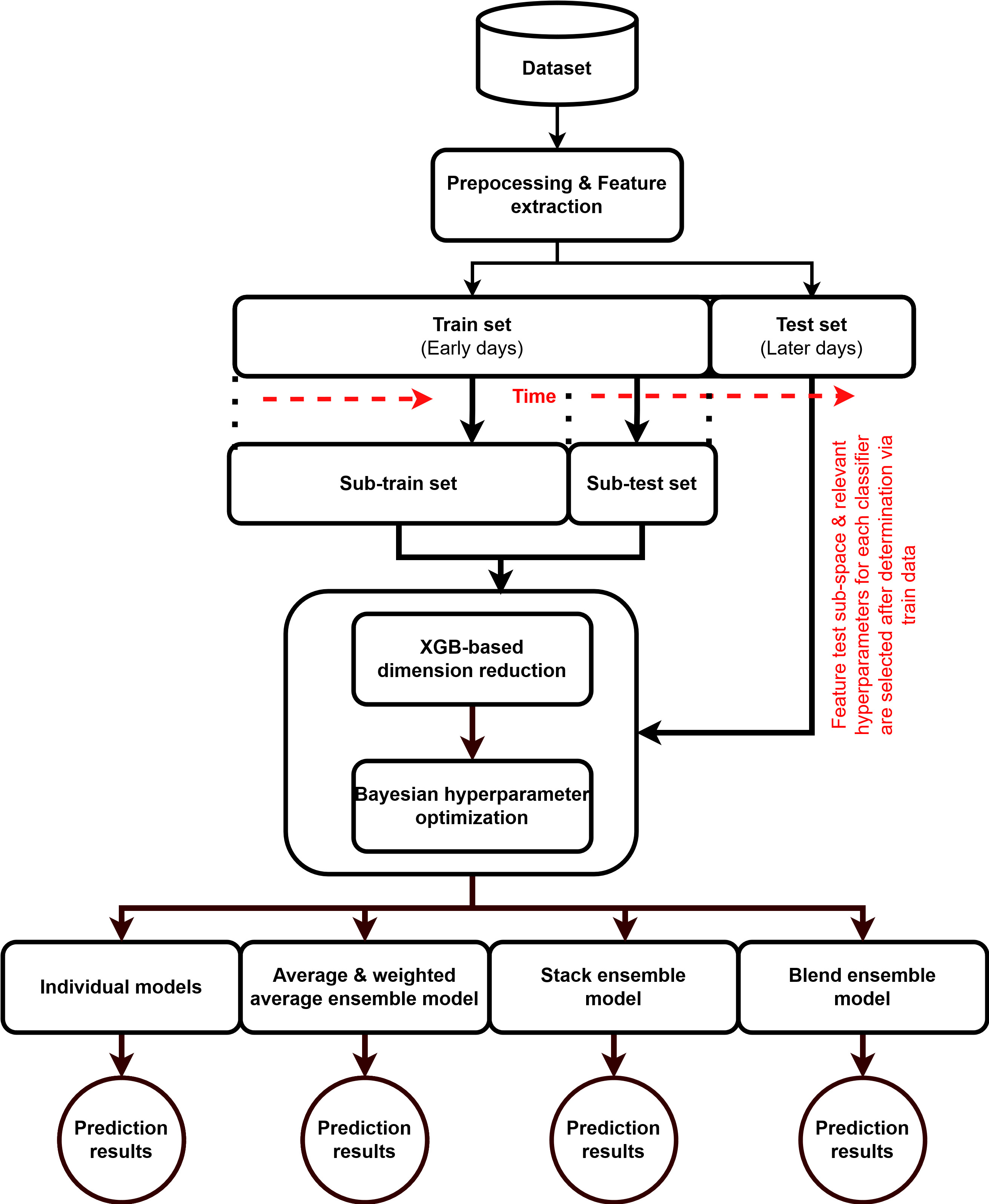}
\caption{Overview of the System. The main train set is divided into two subsets (train and test) to assess the importance of features. These are used to determine the most representative feature subspace by testing with the individual dataset that should be isolated from the actual test set. Accordingly, Bayesian hyperparameter optimization is applied to each individual model via training with a sub-train set. The dimensionality of the main test set is reduced over a predefined feature subspace, and the model is tested over five different model pipelines, including individual ten regressor models, simple averaged and weighted averaged ensemble models, and stack and blend ensemble pipelines.} \label{fig1}
\end{figure}

There are five primary components in the proposed system. The complete system's flow diagram is depicted in Fig.-\ref{fig1}. To summarize, queries are used to retrieve the dataset from the database. Preprocessing is used to extract time-domain seasonal decomposition features with suitable data cleaning in the next stage. XGBoost, LightGBM (LGBM) \cite{bb20} and Stochastic Gradient Descent (SGD) \cite{bb21} algorithms are then subjected to hyper-parameter tuning. In the final step, individual and ensembled models are trained and tested with the same train and test sets to generate click predictions. Each model's $R^2$ score is presented, and 46 distinct models are trained and tested via the proposed system. 


\subsection{Dataset Generation and Data Preprocessing}
The data is retrieved from a major OTA company based on Turkey. Contents of the meta-search platform's daily reports are combined with the data retrieved from the OTA. The dataset contains both numerical and categorical features. Some of the columns are eliminated during the data analysis phase as they contain a high ratio of missing values. In this study, we have replaced the missing values with the most common value and the average of the related feature for categorical and numerical features, respectively.

In addition to OTA's data, some external features are added to the dataset in order to explain the state of the economical and seasonal properties of the environment. Some simple external data examples are daily weather information and daily exchange rates. Data enrichment improves the quality of the dataset. The closeness of the related day to the next public holiday and the length of the holiday are also added as additional numerical variables.

In order to improve the accuracy and generalization ability of the prediction model, additional features are generated from the data following a sliding-window (time-delay) approach. For example, the average and standard deviation of numerical values for some specific time periods (such as the last 3, 7, and 30 days) are calculated and used as input features for prediction. The aim of adding such features is to improve the accuracy and generalization ability of the prediction model. 

Feature space is enriched with the seasonal decomposition of some time-series features. Seasonal decomposition is a naive decomposition model that generates additive components by breaking the original feature into three. The output of the algorithm is T: Trend, S: Seasonality, and e: Residual, where $Y[t] = T[t] + S[t] + e[t]$. The seasonal component is first removed by applying a convolution filter to the data. The average of this smoothed series for each period is the returned seasonal component \cite{bb22}. Decomposed seasonality, trend, and residual values are added to the dataset as new features. 

As a final step, feature one-hot encoding is proposed for some of the string-based features and binarized. In the last step, the feature set is normalized with min-max scaling to force values to be between 0 and 1. 

\subsection{XGBoost-based recursive feature elimination}

XGBoost is the part of gradient boosting decision tree which operate via regularization of the tree framework. By using gradient boosting to create the boosted trees and collect the feature scores in an effective manner, each feature's significance to the training model is indicated \cite{bb22.1}. The formula calculation the feature importance of every feature $F_n$ is shown in Eq. ~\ref{eq:XGB_featureimportance}.

\begin{equation}
F_n(T) = \sqrt{\frac{1}{E} \sum_{e=1}^{E} \hat{i}^{2}(T_e)}
\label{eq:XGB_featureimportance}
\end{equation}

\noindent There is a subdivision of each node into two regions at every node $e$ for each feature $n$ as a part of the feature space $F_n$ from a given single decision tree $T$. The maximally forecasted score boosting rate $\hat{i}^{2}$ represents the metric of squared error shifts of the cost function from the given XGBoost regression outcome of an additive tree $T_e$. The summation of the squared importance over all trees $E$ proposes the summarization of the square importance of the given feature $n$. Accordingly, the root mean squared importance manifests the absolute importance factor of the feature.

The estimation of such an improvement depends on replacing the actual feature value in space with random noise to determine a relative magnitude shift in the final regression performance. Running multiple trees simultaneously provides a better understanding of the average importance of the feature.

In the next step, the customized recursive feature elimination algorithm is used to minimize the feature space \cite{bb22.2}. The algorithm \ref{alg:xgboost_recursive_alg} shows the procedure of the flow. The goal is to cover the features ($feature\_subspace$) that represent best the feature importance levels in a descending order. To avoid the complexity of the classical recursive-based feature elimination due to the large feature space, the initial feature importance values are considered as bias factors for the features. Given that the randomization factor of the selected features will be auto-biased in the subspace, such a specialization significantly reduces the elimination process. $r^2\_score$ value of a new $feature\_subspace$ is calculated within every iteration until convergence occurs ($r^2\_temp$ value stop being exceeded by $r^2\_score$). Again XGBoost regressor is selected as feature sub-space evaluator.

\begin{algorithm}[H]
\caption{Recursive XGBoost dimensionally reduction algorithm}
\label{alg:xgboost_recursive_alg}
\KwData{\\$ \;\;\;\;\;\;\; FI = sort\_descending(feature\_importances)$ \\
        $\;\;\;\;\;\;\; r^2\_temp = 0$}
\KwResult{\\$ \;\;\;\;\;\;\; feature\_subspace$}
 \For{$FI_0 \;\; in \;\; FI$}
  {
    $feature\_subspace = feature\_space\:(FI_0 < FI)$ \\
    $model = initialize\_XGB\_regressor\:()$ \\
    $model = XGB\_regressor\_train\:(train\_data, \; train\_labels)$ \\
    $r^2\_score = XGB\_regressor\_test\:(model, \; test\_data, \; test\_labels)$ \\
    \eIf{$r^2\_score < r^2\_temp$}
    {
     \Return $(feature\_subspace)$
    }{
     $r^2\_temp = r^2\_score$
    }
    
  }    
\end{algorithm}


\subsection{Bayesian Hyper-parameter Optimization}

Hyper-parameter optimization is an essential approach for some machine learning models to enhance prediction performance. There are a few algorithms for tuning hyper-parameters. One of them is a Grid Search \cite{bb23} which tries each combination of given hyper-parameter candidates of a model. Another optimization algorithm is known as random search \cite{bb24}, which randomly extracts hyper-parameter combinations and tries to reach local optima of a performance score. However, none of them are able to reach successful local optima of performance in a short period. Bayesian hyper-parameter optimization \cite{bb25} is a relatively more powerful and efficient algorithm for hyper-parameter tuning. It aims to reach a global optimum in a much shorter time than grid search. There is a probabilistic model of $f(x)$ that aims to be exploited to make decisions about where $X$ is accepted as the next performing function. This procedure helps to find the minimum of non-convex functions in a few epochs, which positively effects the performance. The evaluation metric to rank hyper-parameter combinations through input data is $R-squared (R^2)$. $R-squared$ is a statistical measure that represents the proportion of the variance for a dependent variable that's explained by an independent variable or variables in a regression model. The formula of $R^2$ is shown in Eq.~\ref{eq:R2}.

\begin{equation}
R2 = 1 - \frac{Explained Variation}{Total Variation}
\label{eq:R2}   
\end{equation}


In this work, $R^2$ values of individual machine learning algorithms (XGBoost , LightGBM , SGD, Lasso \cite{bb26}, Lasso Lars \cite{bb27}, Ridge \cite{bb28}, Bayesian Ridge \cite{bb29}, Huber \cite{bb30}, Passive Aggressive Regressors \cite{bb31} and Elastic Net \cite{bb32}) are used and compared in ensemble models.


\subsection{Ensembling}
If there are M models with errors extracted from the same dataset which are uncorrelated with them, the average error of a model is theoretically reduced by some factor by simply averaging the model outputs. On the other hand, if some of the model outputs have lower performance and are not fit to predict results as well as others, overall error may not be reduced or even increase in some cases.


\subsubsection{Average \& Weighted Average of Model Outputs}
The first and most basic ensembling approach is to take an average of various model outputs. There are two different averaging techniques for ensembling. The first one is taking a mean of predicted values. It provides a lower variance of predicted values since different algorithms proceed to predict various aspects of the input data set. The formula for an average of model outputs is shown in Eq.~\ref{eq:avg}.

\begin{equation}
Avg_i = \frac{\sum_{r}^n pi_r}{n}
\label{eq:avg}   
\end{equation}

where $i$ is the $i^{th}$ sample, $r$ is regressor model, $pi_r$ is individual probability of given regressor and $n$ is the number of models used.

However, some machine learning models perform worse than others in terms of prediction, culminating in a poorer overall ensemble prediction performance than some individual regressor prediction performances. The fundamental reason for this is because we give weak regressors the same weight as other ones that provide decent individual performance. As a consequence, while taking an average of all estimations, the weighted average is also utilized in this study to eliminate the detrimental influence of low-performance models. Weights are produced using each model's individual $R^2$ score and scaled between 0 and 1 to standardize the weight of each regressors, ensuring that the sum of all weights is 1.  This method allows models that predict higher performance to have a greater impact on final prediction than models that predict lower performance. The formula of the weighted average of model outputs is shown in Eq.~\ref{eq:wavg}.

\begin{equation}
\begin{split}
Wavg_i = \sum_{r} w_r * pi_r, \\
r \in R \, for \, i= 1 \, to \, N, \\
\sum_{r} w_r = 1     
\end{split}
\label{eq:wavg}    
\end{equation}

where $r$ is the chosen regressor model, $w_r$ is normalized individual $R^2$ performance of regressor. $r$, $pi_r$ is prediction result of regressor $r$ of $i$'th sample and $N$ is the number of models used.


\subsubsection{Stack Ensemble Model}
Stack Ensemble algorithm, assemble results of individual results for different models to make an intermediate input dataset, and the final model is used to create a final regression result. In the proposed approach, ten different models (XGBoost, LGBM, SGD, Lasso, Lasso Lars, Ridge, Bayesian Ridge, Huber, Passive Aggressive Regressors, and Elastic Net) are trained to stack their extracted predictions, and the intermediate dataset, which is the input to ensemble regressors, is also trained with four different meta-regressor models including XGB, Lasso, Bayesian ridge, and linear regressions for the final click predictions. Additional meta-learners are also tried, but due to their immense poor performance, those models are discarded and do not appear in the outcomes of model variants. 

\begin{figure} [H]
\includegraphics[width=0.93\columnwidth]{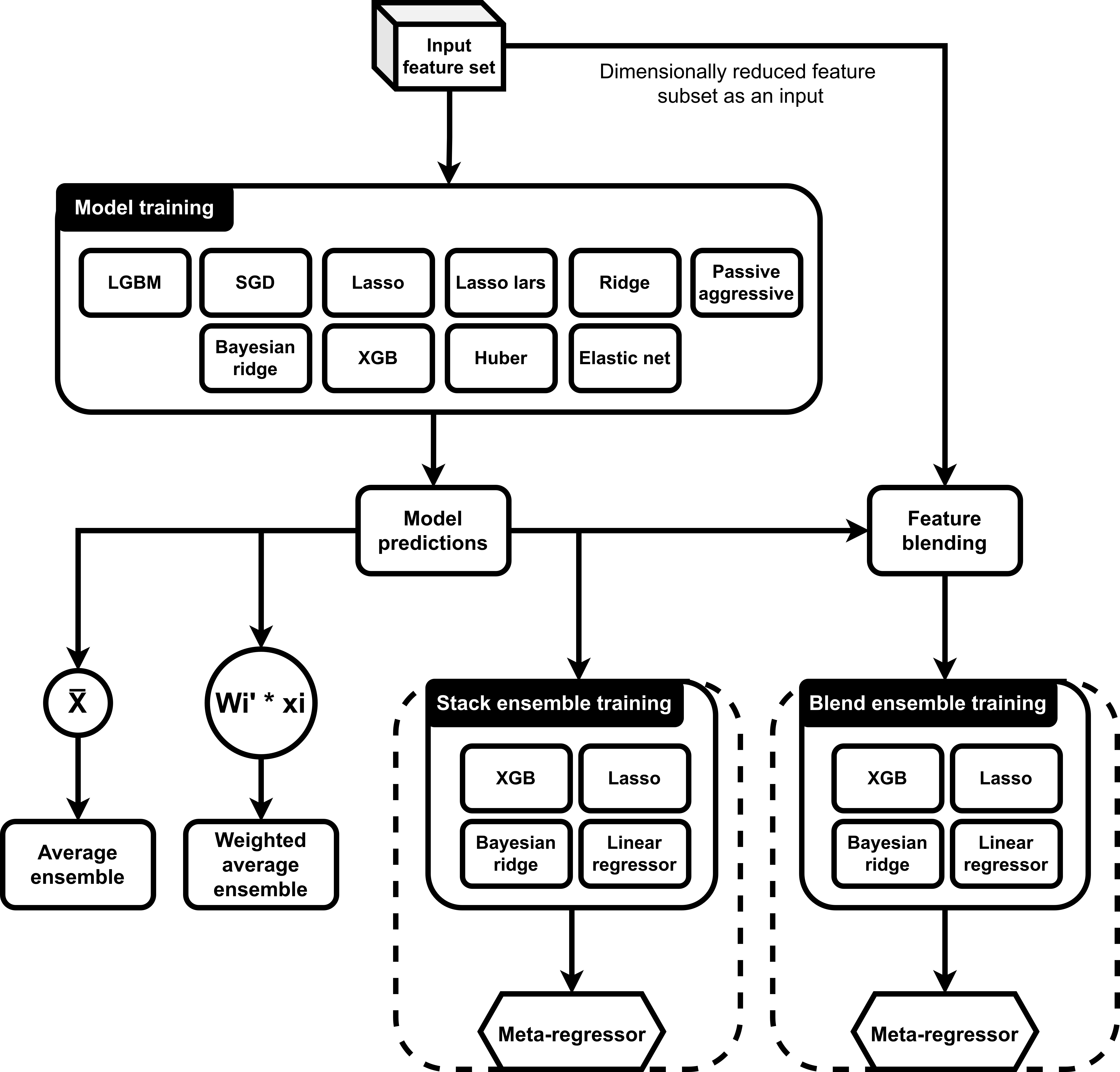}
\caption{Ensemble model pipelines. Individual models are trained via a dimensionally reduced training set. Model predictions are further operated via four ensemble methods: taking the prediction list's average and a weighted average to increase the positive bias for some regressors with better prediction performance; four meta-regressor variations are fed by mediated input features; the blend ensemble learning pipeline via combining a collection of initial feature sets with model prediction results, and feeding the blend into four different meta-regressor variations.} \label{fig:stack_blend_ensemble}
\end{figure}

Stacking the individual predictions enables us to analyze the intermediate regressor model to linearly weight results to create a learnable weighted average of provided predictions through each sample of input data. Overall ensemble model variations are indicated in Fig.~\ref{fig:stack_blend_ensemble} along with the associations between them.

\subsubsection{Blend Ensemble Model}

The Stack ensemble method and the Blend ensemble algorithm \cite{bb34} have similar designs. The separate outcomes of regressor models are assembled in the first stage. Additionally, the individual model outcomes are merged with a dimensionally reduced featureset, which adds mediated features extracted as predicted clicks with knowledge of intended predictions to produce an expanded feature dimension. 

Similar to stack ensemble models, XGBoost, LightGBM, SGD, Lasso, Lasso Lars, Ridge, Bayesian Ridge, Huber, Passive Aggressive Regressors, and Elastic Net are used to stack their given prediction outputs and blended with the input feature set. Then, the blended dataset is also trained with four different models same as the ones (XGB, Lasso, Bayesian ridge, and linear regressions) chosen for the stack ensemble meta-learners to extract four different $R^2$ results. 



\section{Experiments and Results}

Instead of splitting a dataset into train and test with some percentage, daily click predictions of each hotel are estimated. Accordingly, the train set is designed from the earliest day until test day that clicks will be predicted. By using this approach, 11 consecutive days are chosen as test days and 11 corresponding $R^2$ test scores are produced by processing four different ensembling models (Average \& weighted average, stack ensemble, and blend ensemble). Besides, individual $R^2$ test scores of ten regressor models (XGBoost, LightGBM, SGD, Lasso, Lasso Lars, Ridge, Bayesian Ridge, Huber, Passive Aggressive Regressors, and Elastic Net) are reported for the control group, and efficiency of ensemble models is evaluated.

\begin{figure}[H]
\includegraphics[width=0.9\columnwidth]{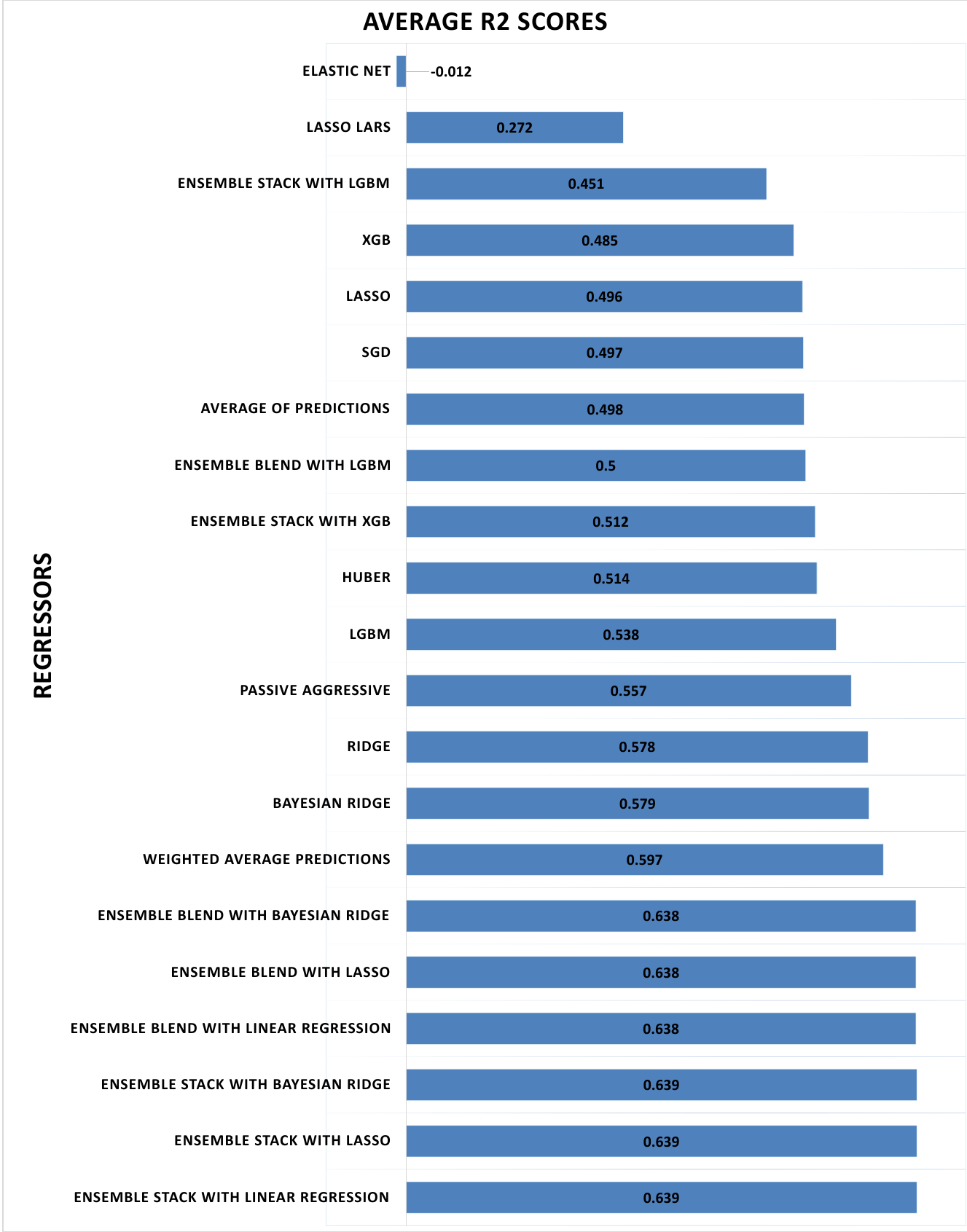}
\caption{Overall Test $R^2$ Scores for Each Regressor Model} \label{fig:testR2Scores}
\end{figure}

For each test day, 22 different predictions are measured (10 individual predictions, average \& weighted average predictions, five stack ensemble prediction, and four blend ensemble predictions). $R^2$ score of each prediction is saved and the average of each test $R^2$ score is calculated. The average $R^2$ test scores of 21 model types are shown in Fig.~\ref{fig:testR2Scores}.

\subsection{Click prediction performances of individual models}

Individual regressor predictions of all models (XGBoost, LightGBM, SGD, Lasso, Lasso Lars, Ridge, Bayesian Ridge, Huber, Passive Aggressive Regressors, and Elastic Net) are reported for the control group, and $R^2$ scores of models are evaluated as 0.485, 0.538, 0.497, 0.496, 0.272, 0.578, 0.579, 0.514, 0.557, and -0.012 respectively.


\subsection{Click prediction performances of ensemble models}

The performance of the ensemble model largely exceeded the results of the individual models, with the highest $R^2$ value of 0.639 shared by three stack ensemble models (ensemble stack with linear regression; ensemble stack with Lasso; and ensemble stack with Bayesian ridge). According to these models, ensemble blending with Lasso and ensemble blending with Bayesian ridge regressors came second at 0.638. The key detail here is that the six best-performing models are ensemble ones. The performance drops relatively significantly on a more primitive ensemble model, the weighted average predictor, which comes third with an $R^2$ value of 0.597. The other three ensemble methods, ensemble stack with XGB, ensemble blend with LGBM, and ensemble stack with LGBM, show performance at the isolated model level (0.512, 0.5, and 0.451).

It can be inferred from the results that simpler regressor models as meta-predictors overshadow tree based regressors due to the fact that the most of the work is already done by the level-0 learners; the level-1 regressor is basically just a mediator and it makes sense to choose a rather simple algorithm for this purpose \cite{bb35}. Simple linear models at the leaves suppose to work well, and the results are likely to prove once again.


\section{Conclusion and discussion}
Assorted regressors are ensembled in the proposed study to improve click prediction performance. The feature set is divided into train and test groups depending on the logging date in the first phase. The data collection is then subjected to an XGBoost-based dimension reduction, which significantly reduces the dimension of features. To discover the most ideal hyper-parameters, Bayesian Hyper-parameter optimization is developed for the XGBoost, LightGBM, and SGD models. XGBoost, LightGBM, SGD, Lasso, Lasso Lars, Ridge, Bayesian Ridge, Huber, Elastic Net, and Passive Aggressive regressors are all tested separately and then fused to create ensemble models. 

The authors suggest four different ensemble approaches. The first ensemble model takes an average of anticipated results as well as a weighted average. A stack ensemble model, for example, assembles all the results of individual forecasts as an intermediate layer that feeds into another individual layer. The third model is a blend ensemble model, which stacks all of the individual prediction outputs and blends them with the original feature set once more. With the outcomes of multiple model outputs, this framework offers an artificial feature generation to boost the feature dimension. 

The same test set is used to test both individual and ensemble models, and the results of 46 model combinations demonstrate that stack ensemble models produce the best $R^2$ score of all. The greatest $R^2$ score is 0.639 for the stack ensemble model combined with linear regression, whereas the best machine learning model had an $R^2$ score of 0.579. As a conclusion, the ensemble model improves prediction performance by about 10\%.  

Various types of artificial neural network (ANN) models will be added to ensemble models in the future, with the goal of improving stack and blend ensemble models. Yandex's CatBoost machine learning model \cite{bb36}, which handles categorical information, can also be added to the list of regressors to examine.

The concept of meta-learners is designed to provide the final outcome, yet there are possibilities to convert them into intermediate learners via inducing additional hyperparameter optimization mechanisms or additional meta-feature elimination due to forming the additive judgement on stacked predictions on an originally reduced feature dimension \cite{bb37}. Articulating meta-learners as mediators would be an inception based regularizer for intercommunication between multiple meta-models as a single pipeline, which might recalibrate incoming feature space with new model parameters to interact with. 

\section{Declaration of Competing Interest}
The authors declare that they have no known competing financial interests or personal relationships that could have appeared to influence the work reported in this paper.



\bibliography{ourBibliography}

\end{document}